\def\year{2022}\relax

\documentclass[letterpaper]{article} 
\usepackage{aaai22}  
\usepackage{times}  
\usepackage{helvet} 
\usepackage{courier}  
\usepackage[hyphens]{url}  
\usepackage{graphicx} 
\usepackage{natbib}  
\usepackage{caption} 
\DeclareCaptionStyle{ruled}{labelfont=normalfont,labelsep=colon,strut=off}
\usepackage{graphicx}
\usepackage{amsmath}
\usepackage{amssymb}
\usepackage{float}

\usepackage{booktabs}       
\usepackage{microtype}      
\usepackage[american]{babel}
\usepackage{graphicx}
\graphicspath{{figure/}}
\usepackage{color}
\usepackage{xcolor}
\definecolor{grassgreen}{rgb}{0.1,0.8,0.2}
\usepackage{threeparttable}

\usepackage{multirow}
\usepackage{subcaption}

\usepackage{lineno}

\def\pt{\phantom{0}}

\newcommand{\squishlist}{
\begin{list}{$\bullet$}
	{ \setlength{\itemsep}{0pt}
		\setlength{\parsep}{1pt}
		\setlength{\topsep}{1pt}
		\setlength{\partopsep}{0pt}
		\setlength{\leftmargin}{1em}
		\setlength{\labelwidth}{1em}
		\setlength{\labelsep}{0.5em} } }

\newcommand{\squishend}{
\end{list}  }

\title{A Random CNN Sees Objects: One Inductive Bias of CNN and Its Applications}

\author{
Yun-Hao Cao, Jianxin Wu\thanks{J. Wu is the corresponding author. This research was supported by the National Natural Science Foundation of China under Grant 61772256 and 61921006.}
}
\affiliations{
State Key Laboratory for Novel Software Technology\\
Nanjing University, China\\
caoyh@lamda.nju.edu.cn, wujx2001@nju.edu.cn
}

\begin{document}

\maketitle

\begin{abstract}
	This paper starts by revealing a surprising finding: without any learning, a randomly initialized CNN can localize objects surprisingly well. That is, a CNN has an inductive bias to naturally focus on objects, named as Tobias (``\underline{T}he \underline{ob}ject \underline{i}s \underline{a}t \underline{s}ight'') in this paper. This empirical inductive bias is further analyzed and successfully applied to self-supervised learning. A CNN is encouraged to learn representations that focus on the foreground object, by transforming every image into various versions with different backgrounds, where the foreground and background separation is guided by Tobias. Experimental results show that the proposed Tobias significantly improves downstream tasks, especially for object detection. This paper also shows that Tobias has consistent improvements on training sets of different sizes, and is more resilient to changes in image augmentations. Code is available at \url{https://github.com/CupidJay/Tobias}.
\end{abstract}

\section{Introduction}

Deep convolutional neural networks (CNNs) have achieved great success in various computer vision tasks. However, as of today we still know little about what makes a CNN suitable for analyzing natural images, i.e., what is its \emph{inductive bias}. The inductive bias of a learning algorithm specifies constraints on the hypothesis space, and a model can only be instantiated from the hypothesis space that satisfies these constraints. It is easy to reveal the inductive bias of certain learning algorithms (e.g., a linear classifier specifies a linear relationship between the features and the target variable). But, the inductive bias of complex CNNs is still hidden in the fog~\citep{bias:cohen:ICLR17}. Successfully identifying CNN's inductive bias will not only deepen our theoretical understanding of this complex model, but also lead to potential important algorithmic progresses.

Objects are the key in most natural images, and CNNs are good at recognizing, detecting and segmenting objects. For instance, weakly supervised object localization (WSOL)~\citep{cam:zhou:CVPR16, gradcam:ICCV17, psol:zhang:CVPR20} and unsupervised object localization (USOL) methods~\citep{scda:tip17, ddt:wei:pr19} can even localize objects without training on bounding box annotations. All these methods, however, \textit{rely on ImageNet~\citep{ILSVRC2012:russakovsky:IJCV15} pretrained models} and \emph{non-trivial learning steps}.

\begin{figure}[t]
	\centering
	\includegraphics[width=\columnwidth]{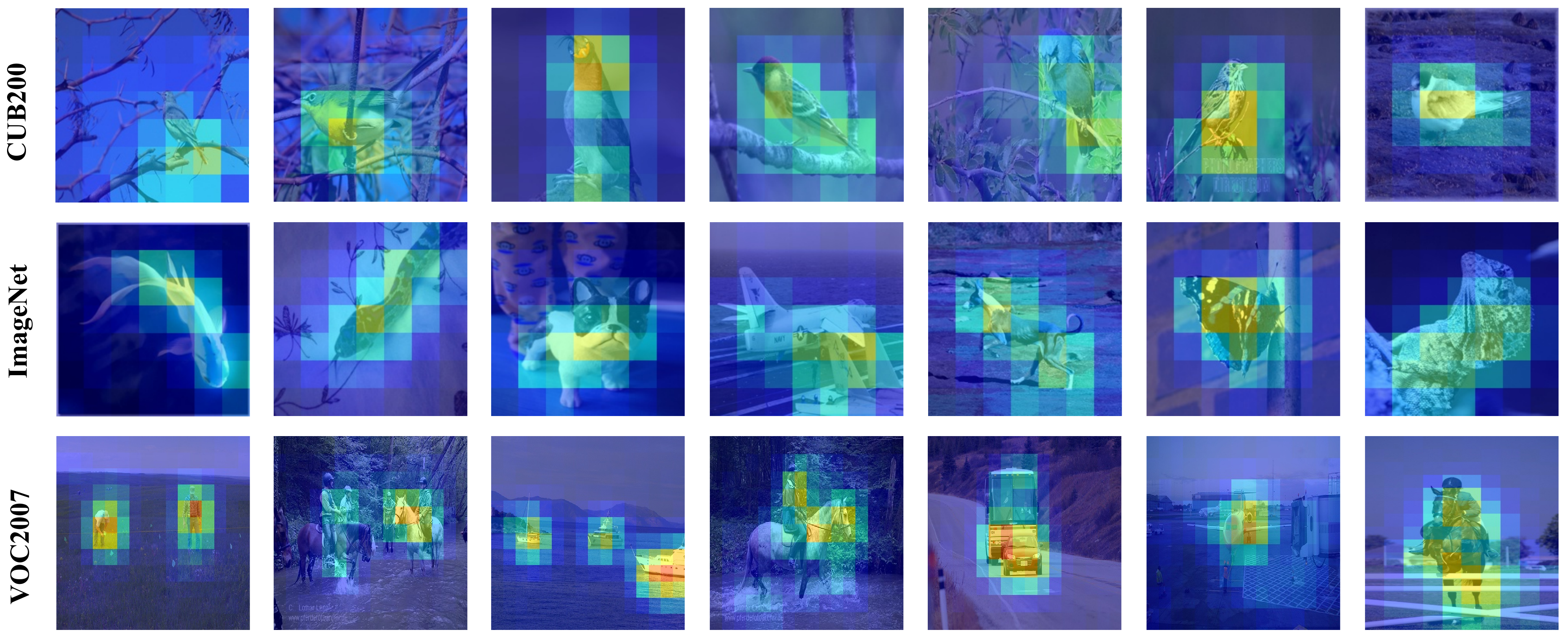}
	\caption{Visualization of localization heatmaps using SCDA~\citep{scda:tip17} for a \textit{randomly initialized} ResNet-50. Best viewed in color when zoomed in.}
	\label{fig:vis}
\end{figure}

In this paper, we first show that focusing its attention to objects is a born gift of CNNs even \emph{without any training}, i.e., it is CNN's inductive bias (or one inductive bias out of many) from an empirical perspective! A \textit{randomly initialized} CNN has surprisingly good localization ability, as shown in Figure~\ref{fig:vis}. We name this phenomenon ``The object is at sight'', or ``Tobias'' for short. The object(s) miraculously pop out (``at sight'') without any need for learning. Our conjecture is: the background is relatively texture-less compared to the objects, and texture-less regions have higher chances to be deactivated by activation functions like ReLU.

Tobias then lends us `free' (free of labels \emph{and} pretrained models) and relatively accurate supervision for where objects are. Hence, a natural application of Tobias is self-supervised learning (SSL), which aims to learn useful representations without requiring labels. After the emerging of the InfoNCE loss~\citep{InfoNCE:arxiv2018} and the contrastive learning paradigm, many SSL algorithms have been published, such as MoCo~\citep{moco:kaiming:CVPR20}, SimCLR~\citep{simclr:hinton:ICML20}, BYOL~\citep{byol:grill:NIPS20}, and many more. In this paper, we propose to probabilistically change an image's background (selected from other images) while keeping the foreground objects by using Tobias. We thus force the model to learn representations focusing on the objects.

We evaluate the representation learned by Tobias SSL on ImageNet and other vision benchmarks. Our method achieves consistent improvements on various benchmarks, especially on object detection because our method can better capture the foreground objects. Also, we carefully study the influence of the number of pretraining images, and our method has consistent improvements on different amounts of training data. Our contributions are: (i) We find the ``Tobias'' inductive bias of CNN, i.e., a random CNN can localize objects without any learning. (ii) We find that activation functions like ReLU and network depth are essential for a random CNN to localize. (iii) We successfully apply Tobias to SSL and achieve consistent improvements on various benchmarks. (iv) Our method is robust when the amount of data is small or large, and is more resilient to changes in the set of image augmentations.

\section{Related Works}

\textbf{Random networks' potential.}~\citet{lottery:frankle:ICLR19} proposed the Lottery Ticket Hypothesis: A randomly initialized, dense neural network contains a subnetwork that is initialized such that---when trained in isolation---it can match the test accuracy of the original network after training for at most the same number of iterations. A lot of works followed this line of research~\citep{lotteryrecognition:girish:CVPR21, lotteryssl:chen:CVPR21, lotteryprove:malach:ICML20}. The SSL method BYOL~\citep{byol:grill:NIPS20} was also motivated by the random network's potential: the representation obtained by using fixed randomly initialized network to produce the targets can already be much better than the initial fixed representation. DIP~\citep{deepimageprior:Dmitry:CVPR18} proposed that a randomly initialized neural network can be used as a handcrafted prior in standard inverse problems. These works show the potential of random networks from the perspective of network pruning, self learning or image denoising. We investigate it from a new perspective: a random CNN sees objects. 

\textbf{Un/Weakly-supervised object localization.} Weakly supervised object localization (WSOL)~\citep{gradcam:ICCV17,psol:zhang:CVPR20} learns to localize objects with only image-level labels. CAM~\citep{cam:zhou:CVPR16} generated class activation maps with the global average pooling (GAP) layer and the final fully connected (FC) layer (weights of the classifier). Unsupervised localization methods do not even need image-level labels. SCDA~\citep{scda:tip17} aggregated information through the channel dimension to get localization masks. DDT~\citep{ddt:wei:pr19} evaluated the correlation of descriptors. However, they all rely on ImageNet~\citep{ILSVRC2012:russakovsky:IJCV15} pretrained models. Instead, our Tobias does not require any labels or pretrained models.

\textbf{Self-supervised learning.} Self-supervised learning (SSL) has emerged as a powerful method to learn visual representations without the expensive labels. Many recent works follow the contrastive learning paradigm~\citep{InfoNCE:arxiv2018}. SimCLR~\citep{simclr:hinton:ICML20} and MoCo~\citep{moco:kaiming:CVPR20} trained networks to identify a pair of views originating from the same image when contrasted with a large set of views from other images. The most related methods to ours are~\citep{unmix:shen:arxiv20} and~\citep{bsim:cgy:arxiv20}, where Mixup~\citep{mixup:ICLR18} or CutMix~\citep{cutmix:yun:ICCV19} was used to combine two images and force the new image to be similar to both. However, they may either generate unnatural images or cut objects out due to the lack of supervision. In contrast, our method provides free foreground vs. background supervision to merge patches, which proves to be useful in subsequent experiments.

\textbf{Data augmentation.} We use Tobias to merge patches from two different images to generate a new image, which keeps the objects and replaces the background. Our method can be viewed as a data augmentation strategy. As aforementioned, Mixup and CutMix do not have the location information as in our method and the random cut in CutMix may cover the foreground area with the background. ``Copy and paste''~\citep{cut-paste-learn:dwibedi:ICCV19, simple-copy-paste:quoc:arxiv2012} is an effective augmentation in object detection and instance segmentation, which cut object instances and paste them on other images. These methods require ground-truth bounding box labels, while ours does not rely on any labels. 

\section{Tobias, and SSL with Tobias} \label{sec:method}

Now we first introduce how a randomly initialized CNN localizes objects. Then, we introduce how Tobias is applied to self-supervised learning.

\subsection{Object localization using a random CNN} \label{sec:method1}

Given an input image $\boldsymbol{x}$ of size $H\times{W}$, the outputs of a CNN (before the GAP layer) are formulated as an order-3 tensor $Q\in{\mathbb{R}^{h\times{w}\times{d} }}$, which include a set of 2-D feature maps $S=\{S_n\} (n=1,\dots,d)$. $S_n$ (of size $h\times{w}$) is the $n$-th feature map of the corresponding channel (the $n$-th channel). For instance, by employing the ResNet-50~\citep{resnet:he:CVPR16} model, $Q$ is the output of `pool5' (i.e., activations of the last max-pooling layer) and we can get a $7\times7\times2048$ tensor if the input image is $224\times224$.

SCDA~\citep{scda:tip17} obtains a 2-D aggregation map $A\in{\mathbb{R}^{h\times{w}}}$ by adding up $Q$ through the depth direction and then uses the mean value of $A$ as the threshold to localize objects. Formally, $A=\sum_{n=1}^d{S_n}$. Then, a mask map $M$ of the same size as $A$ can be obtained by
\begin{equation}
	M_{i,j} = \left\{
	\begin{array}{rcl}
		1&  & {\text{if}\ A_{i,j} > \bar{a}}\\
		0 && {\text{otherwise}}
	\end{array} \right. ,
\end{equation}
where $\bar{a}=\frac{1}{h\times{w}}\sum_{i,j}A_{i,j}$ and $(i,j)$ denotes any position in these $h\times{w}$ locations. Those positions $(i,j)$ whose activation responses are higher than $\bar{a}$ (i.e., $M_{i,j}=1$) indicate the foreground objects.

The original SCDA~\citep{scda:tip17} used ImageNet pretrained models for feature extraction and localization, and obtained good localization performance. However, there are many scenarios where pretrained models do not exist. Instead, we follow the same setups as in SCDA but replace the ImageNet pretrained weights by random weights. We find that a pretrained model is not necessary and a randomly initialized CNN can also localize objects surprisingly well. We name this phenomenon ``\underline{T}he \underline{ob}ject \underline{i}s at \underline{s}ight'', or ``Tobias'' for short. Figure~\ref{fig:vis} visualizes some localization examples, and we defer more results and analyses to the next section.

\begin{figure}[t]
	\centering
	\includegraphics[width=0.95\columnwidth]{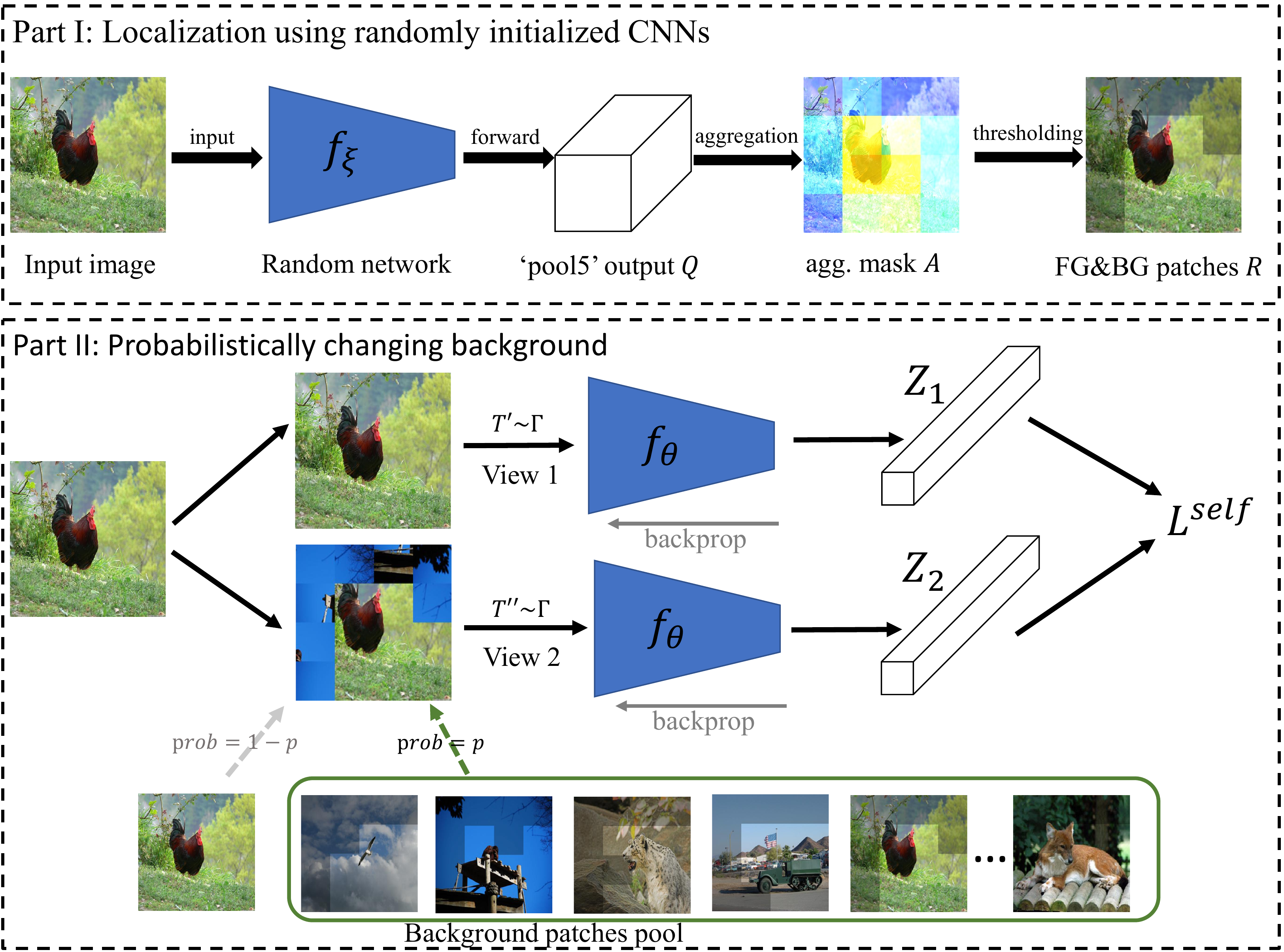}
	\caption{Pipeline of Tobias SSL. Upper part: splitting foreground and background using a \textit{randomly initialized} CNN. Lower part: applying Tobias augmentation into SSL.}
	\label{fig:network}
\end{figure}

\subsection{Tobias self-supervised learning} \label{sec:method2}

Based on our finding that an un-trained random network can capture foreground objects surprisingly well (i.e., Tobias), it is natural to wonder if we can take advantage of this property in SSL, where we do not have any pretrained models or annotated labels. In this section, we propose a Tobias augmentation, which keeps the objects and probabilistically changes the background for an image, and can be integrated into any existing SSL method. 
Moreover, we will demonstrate that our method can be viewed as either a data augmentation or a pseudo supervised contrastive learning method.

\textbf{The Tobias augmentation.} We make two modifications to SCDA in order to better adapt to SSL algorithms. First, we add an extra max-pooling layer (with stride=2) after `pool5' and the mask map $M$ becomes $4\times4$ instead of $7\times7$ for a $224\times224$ input image. The mask $M$ for each image is pre-calculated by a randomly initialized network and do not change during further training. Second, we use \textit{the median} instead of the mean value as the threshold to make sure that we have half the background ($M_{i,j}=0$) and half the foreground ($M_{i,j}=1$). Notice that this hard half-half division cannot fit all images exactly, because there exist images where objects cover more than or less than half of the area. However, this choice makes it easier when we combine foreground and background patches from two different images.

Then we split the input image $\boldsymbol{x}$ into $4\times4=16$ patches $R=\{R_{i,j}\}(i,j=0,\dots,3)$, in which each patch corresponds to one position in $M$:
\begin{equation}
	R_{i,j} = \boldsymbol{x}[i \times r:(i+1) \times r-1,j \times r:(j+1) \times r-1] \,,
\end{equation}
where $[:,:]$ denotes the slice operation, $r\times{r}$ is the patch size and $r=224/4=56$ in our setting. We call $R_{i,j}$ a foreground patch if $M_{i,j}=1$ and a background patch otherwise.

Given two image $\boldsymbol{x}_1$ and $\boldsymbol{x}_2$, we can generate a new image $\boldsymbol{x}_{1,2}$, which contains foreground patches in $\boldsymbol{x}_1$ and background patches in $\boldsymbol{x}_2$. When merging patches from two images, we keep the positions of foreground patches unchanged and fill in other positions with background patches in a random order. Let $R^{(1)},R^{(2)}$ and $R^{(1,2)}$ denote the patches in $\boldsymbol{x}_1, \boldsymbol{x}_2$ and $\boldsymbol{x}_{1,2}$, respectively. Then,
\begin{equation}
	\label{Tk}
	R^{(1,2)}_{i,j} = \left\{
	\begin{array}{rcl}
		R^{(1)}_{i,j}\phantom{0} &  & \text{if} \  M^{(1)}_{i,j}=1\\
		R^{(2)}_{\sigma(i,j)} && {\text{otherwise}}
	\end{array} \right. ,
\end{equation}
where $\sigma(\cdot,\cdot)$ defines a one-to-one mapping from background positions in $\boldsymbol{x}_1$ to background positions in $\boldsymbol{x}_2$. More specifically, background positions in $\boldsymbol{x}$ means $\{(i,j)|M_{i,j}=0\}$ and $\sigma$ defines such a random order to fill in background patches. Notice that all images have the same number of foreground and background patches and we are safe and free to merge these patches.

\textbf{Applying Tobias to SSL.} We now apply Tobias to the contrastive learning paradigm following the notations in SupCon~\citep{supcon:khosla:nips20}. Suppose the dataset $D$ has a total of $N_t$ images and we randomly sample $N$ images $\{\boldsymbol{x}_k\}_{k=1\dots{N}}$ to form a batch. The corresponding batch used for training consists of $2N$ pairs, $\{\boldsymbol{x}'_k,\boldsymbol{x}''_k\}_{k=1\dots{N}}$, where $\boldsymbol{x}'_{k}$ and $\boldsymbol{x}''_{k}$ are two random augmentations (i.e., ``views'') of $\boldsymbol{x}_{k}$. We denote the transformation as $T$, which is sampled from the predefined augmentation function space $\Gamma$. Hence we have $\boldsymbol{x}'_{k}=T'(\boldsymbol{x}_k)$ and $\boldsymbol{x}''_{k}=T''(\boldsymbol{x}_k)$, where $T',T''\sim\Gamma$. In self-supervised contrastive learning, e.g., MoCo~\citep{moco:kaiming:CVPR20}, the loss takes the following form:
\begin{equation}
	L^{self}=-\sum_{i} \log \frac{e^{\boldsymbol{z}'_i\boldsymbol{\cdot}\boldsymbol{z}''_i/\tau}}{\sum_{j\neq i} e^{\boldsymbol{z}'_i\boldsymbol{\cdot} \boldsymbol{z}'_j/\tau}+\sum_{j} e^{\boldsymbol{z}'_i\boldsymbol{\cdot} \boldsymbol{z}''_j/\tau}},
\end{equation}
where $\boldsymbol{z}'_i=f(\boldsymbol{x}'_i)$, $\boldsymbol{z}''_i=f(\boldsymbol{x}''_i)$, the $\boldsymbol{\cdot}$ symbol denotes the inner product and $\tau$ is the temperature parameter. Here $f(\cdot)\equiv\text{Proj}(\text{Enc}(\cdot))$ denotes the composition of a encoder and a projection network.

Then we introduce Tobias into SSL (illustrated in Figure~\ref{fig:network}). Given an image $\boldsymbol{x}_k$, we generate the first view as before, i.e., $\boldsymbol{x}'_k=T'(\boldsymbol{x}_k)$. However, for another view $\boldsymbol{x}''_k$, we transform $\boldsymbol{x}_k$ into $\boldsymbol{x}_{k,m}$ by changing its background patches with another randomly selected image $\boldsymbol{x}_m$ with probability $p$, where $p$ is a hyper-parameter:
\begin{equation}
	\label{view2}
	\left\{
	\begin{array}{lcl}
		Pr\big(\boldsymbol{x}''_k=T''(\boldsymbol{x}_k)\big)=1-p&\\
		Pr\big(\boldsymbol{x}''_k=T''(\boldsymbol{x}_{k,m})\big)=\frac{p}{N_t}&, m=1,\dots,N_t
	\end{array} \right. .
\end{equation}

Hence, the loss function becomes
\begin{equation}
	\label{pcb}
	L^{Tobias}=-\sum_{i} \log \frac{e^{\boldsymbol{z}'_i\boldsymbol{\cdot}\boldsymbol{z}^p_i/\tau}}{\sum_{j\neq i} e^{\boldsymbol{z}'_i\boldsymbol{\cdot} \boldsymbol{z}'_j/\tau}+\sum_{j} e^{\boldsymbol{z}'_i\boldsymbol{\cdot} \boldsymbol{z}^p_j/\tau}},
\end{equation}
where $\boldsymbol{z}^p_i=f(\boldsymbol{x}^p_i)$ and $\boldsymbol{x}^p_i$ is one of the augmented samples in $P(i)\equiv\{\boldsymbol{x}_i, \boldsymbol{x}_{i,1}, \dots, \boldsymbol{x}_{i,N_t}\}$, which follows the distribution in Equation~\ref{view2}. Notice that when $p=0$, $L^{Tobias}$ degenerates into $L^{self}$. Furthermore, Equation~\ref{pcb} can be seen as a pseudo supervised contrastive loss, where $P(i)$ contains images with the same foreground object.

\begin{table*}[!htbp]
	\caption{Comparisons of localization accuracy between ImageNet pretrained and randomly initialized CNNs on ImageNet and CUB-200. `\#ReLU' and `\#stages' represent the number of ReLU units and stages, respectively. `IN super.' stands for `ImageNet supervised'. We report the average accuracy and standard deviation of 3 trials for randomly initialized models.}
	\label{tab:loc-results}
	\centering
	\renewcommand{\arraystretch}{0.9}
	\renewcommand{\multirowsetup}{\centering}
	\begin{tabular}{l|l|c|c|c|c|c}
		\hline
		\multirow{2}{*}{Method}   & \multirow{2}{*}{Backbone}  &
		\multirow{2}{*}{\#ReLU / \#stages} &
		\multicolumn{2}{c|}{ImageNet} & \multicolumn{2}{c}{CUB-200}  \\
		\cline{4-7}
		&&&IN super.&random init.&IN super.&random init.\\
		\hline
		\multirow{8}{*}{SCDA~\citep{scda:tip17}}&R-50& 33 / 5&51.9&48.2$\pm$0.6&44.8&41.8$\pm$0.6\\
		&R-50 (sigmoid)& \pt0 / 5 &46.9&45.5$\pm$1.9  &32.6 &22.6$\pm$3.3\\
		&R-50 (arctan)& \pt0 / 5 &34.4&36.6$\pm$0.7  &19.1 &18.1$\pm$0.3\\
		&R-50 (conv1)& \pt1 / 1 &44.1&41.3$\pm$1.5  &33.8 &30.5$\pm$1.0\\
		&R-50 (conv1-2)& \pt7 / 2&38.4&39.7$\pm$1.5  &22.1 &29.6$\pm$0.9\\
		&R-50 (conv1-3)&15 / 3 &45.0& 42.2$\pm$0.9 &31.0 &31.8$\pm$0.2\\
		&R-50 (conv1-4)& 27 / 4 &49.9&47.2$\pm$1.3  &39.2 &40.1$\pm$0.4\\
		\cline{2-7}
		&Vit-Base & - / - & 50.9 & 40.5$\pm$0.5 & 48.6 & 31.9$\pm$1.3 \\		
		\hline
		CAM~\citep{cam:zhou:CVPR16} &R-50&33 / 5&52.9&33.8$\pm$0.1&50.0 &26.0$\pm$0.3\\
		\hline
	\end{tabular}
\end{table*}

\section{Experimental Results} \label{sec:exp}

We use CUB-200~\citep{cub200} and ImageNet~\citep{ILSVRC2012:russakovsky:IJCV15} for our experiments. First, we show the localization results of randomly initialized CNNs and make further analyses. Then, we apply our Tobias method into SSL and demonstrate its effectiveness across various pretraining datasets, downstream tasks, backbone architectures and SSL algorithms. Finally, we study the effects of different components and hyper-parameters and sensitivity to data augmentations in our algorithm. All our experiments were conducted using PyTorch~\citep{pytorch:NIPS19} and we used 8 Titan Xp GPUs for our experiments.

\subsection{Localization ability of random CNNs} \label{sec:exp1}

In this section, we study the localization ability of randomly initialized CNNs. We use SCDA~\citep{scda:tip17} for localization and conduct experiments on two popular datasets for object localization, i.e., ImageNet and CUB-200. Notice that \citet{scda:tip17} used ImageNet pretrained models for evaluation while we study the potential of randomly initialized models here. The localization is correct when the intersection over union (IoU) between the ground truth bounding box and the predicted box is 50\% or more. In Table~\ref{tab:loc-results}, we report the average localization accuracy and standard deviation of 3 trials for randomly initialized models and we adopt Kaiming initialization~\citep{kaiminginit:he:ICCV15} used in the PyTorch official code. We use the PyTorch official models for ImageNet pretrained models. We show some visualization results on CUB-200, ImageNet as well as one complex multi-object dataset Pascal VOC2007~\citep{VOC:mark:IJCV10} in Figure~\ref{fig:vis}. The heatmap in Figure~\ref{fig:vis} is calculated by the 2-D aggregation mask $A$, as noted in the previous section.

As shown in Table~\ref{tab:loc-results}, a randomly initialized ResNet-50 (R-50)~\citep{resnet:he:CVPR16} achieves comparable localization accuracies with its ImageNet supervised counterpart on both ImageNet and CUB-200. We also present one popular WSOL method CAM~\citep{cam:zhou:CVPR16} for comparison and it further shows that our results for random CNNs are accurate. Notice that SCDA relies only on convolution feature maps while CAM also relies on the trained FC weights, hence we can see a significant drop for CAM with randomly initialized models. Also, from Figure~\ref{fig:vis} we can observe more intuitively that randomly initialized CNNs can not only locate a single object, but multiple objects as well. Furthermore, we can observe that the standard deviation of multiple trials is small for randomly initialized models (there is also only small difference between the visualization results of different trials). The results show that a randomly initialized CNN can achieve surprisingly good localization results and the localization results are robust with different random weights. Moreover, as the core component of CNNs is convolution, we also investigate what the localization effect has to do with convolution. We compare with the non-CNN architecture ViT-Base~\citep{vit:ICLR21} and there is a large gap between the pretrained and randomly initialized ViT models. Hence, we can conclude that it is one inductive bias for CNNs, not for MLP-based architectures, e.g., ViT. 

But, why can a random CNN see objects without any learning? Given the empirical results and in particular its stability under different random initializations, we believe it is the inductive bias of modern CNNs. There are a lot of ReLU and convolution layers inside ResNet-50 (and most other modern CNNs). Remember that SCDA simply adds feature maps across the channel dimension. Hence, if one spatial location has many zeros (i.e., deactivated after ReLU), we expect it to have a low SCDA score and thus being predicted as belonging to the background. 

Our conjecture is then: \textit{the background is relatively texture-less when compared to the objects, and texture-less regions have higher chances to be deactivated by ReLU when the network depth increases.} We design two experiments to verify it. One is to replace all ReLU activations with other activation functions (e.g., sigmoid). The other one is to gradually reduce the number of ReLU units and we directly remove whole stages for R-50. For instance, `conv1-4' means that we remove the last stage in R-50 (i.e., `conv5'). From Table~\ref{tab:loc-results} we can have the following two conclusions. First, ReLU plays an important role because when we replace ReLU with sigmoid or arctan, a significant decrease in localization accuracy was observed. Second, network depth is also important and we can observe a significant performance degradation as the network depth decreases (i.e., fewer stages). 

\begin{table}
	\caption{Localization accuracy of various CNNs on ImageNet and CUB-200. We report the average accuracy and standard deviation of 3 trials for randomly initialized models.}
	\label{tab:loc-results-more}
	\centering
	\setlength{\tabcolsep}{3pt}
	\renewcommand{\arraystretch}{0.95}
	\renewcommand{\multirowsetup}{\centering}
	\begin{tabular}{r|l|c|c}
		\hline
		id &Backbone  &
		ImageNet & CUB-200 \\
		\hline
		1&R-50 &48.2$\pm$0.6	&41.8$\pm$0.6 \\
		2&R-50 (w/o skip connection) & 50.8$\pm$1.0&42.3$\pm$1.5 \\
		3&R-50 (w/o batch normalization) &49.1$\pm$0.5&41.0$\pm$1.4 \\
		4&R-50 (shallow [1,2,3,1])&43.9$\pm$1.5&36.7$\pm$0.7 \\
		5&R-50 (shallow [1,1,1,1])&42.1$\pm$1.7&	31.8$\pm$2.1\\
		6&R-50 (deep [6,8,12,6])&50.0$\pm$0.8&45.0$\pm$0.9\\
		7&R-50 (ELU)&	49.3$\pm$0.9 & 46.6$\pm$2.7\\
		8&R-50 (SELU)& 50.4$\pm$0.6 & 45.4$\pm$3.5\\
		9&R-50 (softplus)& 51.0$\pm$2.7&	51.6$\pm$3.1\\
		10&R-50 (init: Normal(0,0.1)) & 50.6$\pm$0.3 & 43.4$\pm$0.5\\
		11&R-50 (init: Uniform(-0.1,0.1)) & 50.0$\pm$0.4 & 43.4$\pm$0.4 \\
		12&R-50 (init: Xavier) & 42.2$\pm$1.1 & 32.6$\pm$0.9 \\
		\hline
		13&VGG-11 &	40.0$\pm$0.5&	30.6$\pm$1.4 \\
		14&VGG-16 &	40.8$\pm$0.5&	33.5$\pm$1.9\\
		15&VGG-16 (sigmoid)&	39.8$\pm$0.6&	18.2$\pm$1.3 \\
		16&VGG-16 (arctan)	&34.6$\pm$0.5&20.1$\pm$1.0\\
		17&VGG-16 (ELU)	&40.4$\pm$1.0&32.5$\pm$1.0\\
		18&VGG-19 &	41.4$\pm$1.8&	34.2$\pm$0.3\\
		\hline
		19&AlexNet	&34.6$\pm$1.5&	24.8$\pm$0.3 \\
		\hline
		20&Inception v3&	52.2$\pm$0.6&49.6$\pm$0.9\\
		\hline
		21&Hourglass& 52.6$\pm$0.2&46.9$\pm$0.4\\
		\hline
		22&EdgeBox&	31.8&	32.7\\
		23&lower bound (whole image)&	38.8&	19.1\\
		24&upper bound (faster R-CNN)&	58.9&	96.2\\
		\hline
	\end{tabular}
\end{table}

To make our conclusions more convincing, we add more baselines and further investigate different components in ResNet-50 as well as other CNN architectures in Table~\ref{tab:loc-results-more}.

\noindent\textbf{More baselines. }We provide some more upper and lower bounds to help understanding. The lower bound is the accuracy of predicting the entire image as the bounding box (not trivial given that many images have a single prominent object in CUB and ImageNet). The upper bound is the accuracy of pretrained Faster R-CNN R-50~\citep{faster-rcnn:ren:NIPS15}, which is directly supervised on the detection task COCO~\citep{coco:LinTY:ECCV14}. We also compare with object proposal method EdgeBox~\citep{edgebox:pitor:ECCV14}. These results further prove that the localization results of random CNNs are good.

\noindent\textbf{Different components in ResNet-50.}

(1) Skip connection and batch normalization (BN)~\citep{batchnorm:Ioffe:ICML15} are not crucial. We remove all skip connections in R-50 (row 2) and we even achieve slightly better performance than the original R-50 (row 1). Also, when we remove all BN (row 3), we achieve comparable performance.

(2) Network depth is important. We reduced the number of stages in Table~\ref{tab:loc-results} before and now we keep the number of stages unchanged but change the number of bottlenecks in each stage. The number of bottlenecks in each stage for R-50 is 3, 4, 6 and 3, respectively (denoted as [3,4,6,3]). When reduced to [1,2,3,1] (row 4), we have 4.3 and 5.1 points decrease compared with original R-50 on ImageNet and CUB, respectively. When further reduced to [1,1,1,1] (row 5), we have 6.1 and 10.0 points decrease on ImageNet and CUB, respectively. Conversely, when we increase the number of bottlenecks to [6,8,12,6], we can get 1.8 and 3.2 points gains on ImageNet and CUB, respectively. It indicates that deeper architectures can better capture high-level information and localize objects better, even when randomly initialized.

(3) Other ReLU-like unbounded activations also help. When we use other activations, e.g., ELU, SELU and softplus (row 7$\sim$9), we can get comparable or even better results than ReLU. All these activations have one thing in common: deactivate negative values and unbounded for positive values. 

\begin{figure}[t]
	\centering
	\includegraphics[width=\columnwidth]{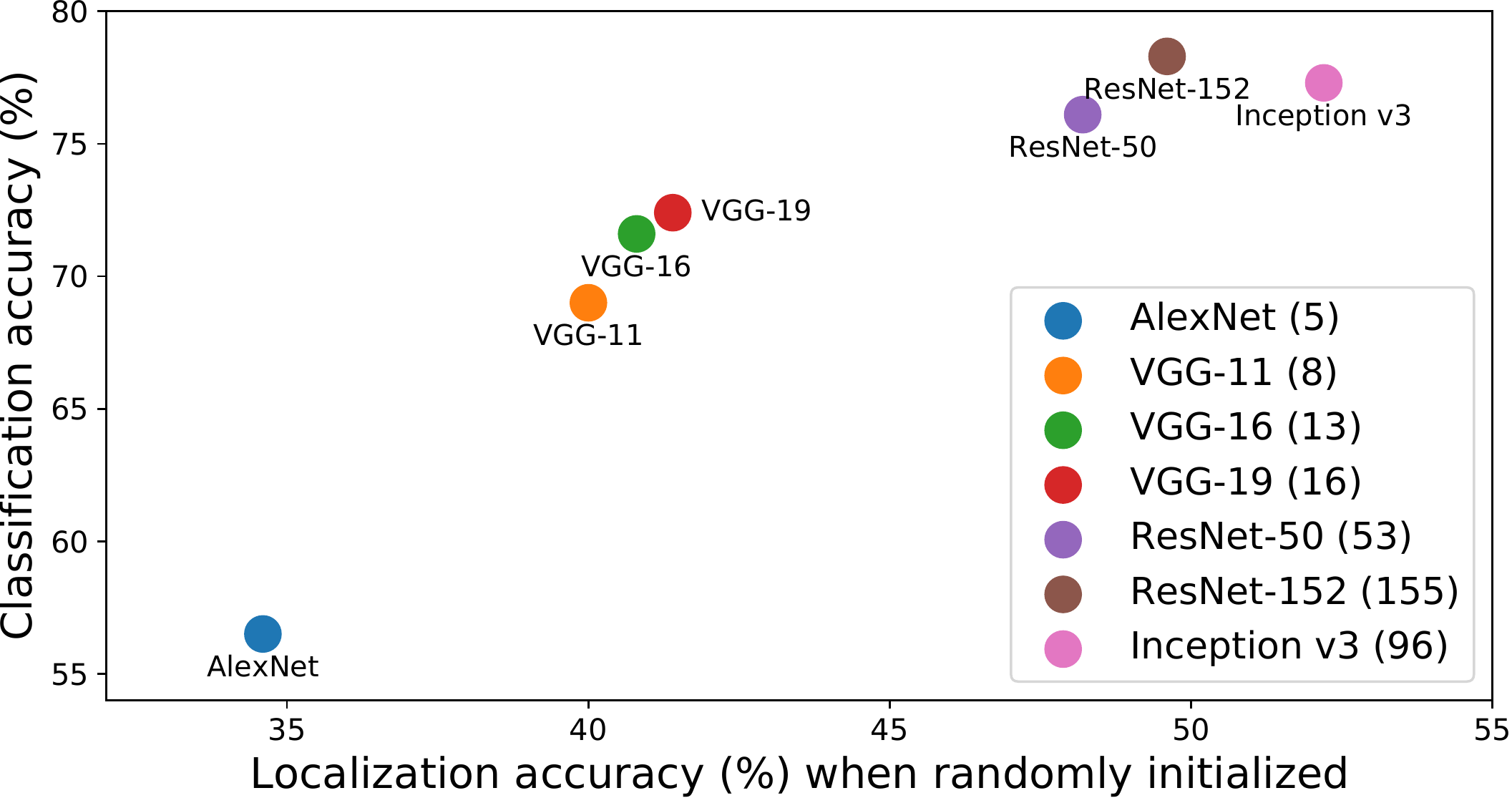}
	\caption{Classification accuracy after training (PyTorch model zoo) versus localization accuracy (when randomly initialized) on ImageNet. The number in brackets represents the number of convolutions in the model (i.e., depth).}
	\label{fig:scatter}
\end{figure}

\noindent\textbf{Other CNN architectures.} Other randomly initialized CNN architectures can also localize objects well. We study with AlexNet~\citep{alexnet:NIPS2012}, VGG-style networks~\citep{vgg:simonyan:ICLR15}, Hourglass network~\citep{deepimageprior:Dmitry:CVPR18} and Inception v3~\citep{inceptionv3:Szegedy:CVPR16}. We can observe that other architectures (e.g., VGG-19, Hourglass and Inception v3) can also achieve non-trivial localization ability. 

When comparing among VGG-style networks, we can also observe that the localization accuracy increases with the increase of network depth (row 13$\sim$18). Also, activations like ReLU perform better than sigmoid and arctan activations. 

When comparing among different CNN architectures, we can see that deeper networks (e.g., Inception v3) perform better than shallow networks (e.g., AlexNet). As shown in Fig.~\ref{fig:scatter}, we rank the following according to the localization accuracy on ImageNet: AlexNet$<$VGG-11$<$VGG-16$<$VGG-19$<$ResNet-50$<$Inception v3. Note that the ranking of the localization accuracy of these random networks is surprisingly consistent with their classification accuracy on ImageNet. We can conclude that deeper networks perform better in terms of localization, even when randomly initialized.

\noindent\textbf{Initialization scheme.} We observe that other initialization methods can also get good and robust localization results (row 10 and row 11) and Tobias is actually a general phenomenon. However, as discussed by~\citet{kaiminginit:he:ICCV15}, the linear assumption in Xavier initialization~\cite{xavier} is invalid for ReLU and we can observe degraded localization accuracy (row 12) here.  

In short, we find that randomly initialized CNNs can localize objects surprisingly well, which is even comparable to their supervised counterparts. Also, we analyze the effect of different components in modern CNNs. The results reveal the potential of a random CNN in localizing objects and provide a new perspective to explain why modern CNNs achieve such good performance in visual analysis.

\subsection{Tobias self-supervised learning} \label{sec:exp2}

Now we apply Tobias to SSL (Equation~\ref{pcb}) and evaluate its effectiveness on CUB200 and ImageNet. Then, we will analyze the effects of different components and hyper-parameters and the sensitivity to data augmentations.

\subsubsection{Results on CUB-200} \label{sec:cub}

\begin{table}
	\caption{Comparisons of pretraining details and accuracies (\%) on CUB-200. `N/A' means that pretraining are conducted on ImageNet instead of CUB-200 for ImageNet supervised models. `FT' is short for `fine-tuning'.}
	\label{tab:clean-cub200-result}
	\centering
	\renewcommand{\arraystretch}{0.95}
	\setlength{\tabcolsep}{3pt}
	\small
	\renewcommand{\multirowsetup}{\centering}
	\begin{tabular}{l|l|c|c|c}
		\hline
		\multirow{2}{*}{Backbone}      & \multicolumn{2}{c|}{SSL pretraining} & \multicolumn{2}{c}{Fine-tuning accuracy (\%)}  \\
		\cline{2-5}
		&method&epochs&Normal FT&Mixup FT\\
		\hline
		\multirow{10}{*}{ResNet-18}& ImageNet super.&N/A &    76.2 \pt\pt\pt\pt\pt &75.0 \pt\pt\pt\pt\pt\\
		\cline{2-5}
		& random init. &0& 62.0 \pt\pt\pt\pt\pt&63.4 \pt\pt\pt\pt\pt \\
		\cline{2-5}
		& MoCov2 & \multirow{2}{*}{200}     &63.7 \pt\pt\pt\pt\pt&65.8 \pt\pt\pt\pt\pt\\
		&MoCov2-Tobias&&64.4 (\textcolor{grassgreen}{+0.7})&66.3 (\textcolor{grassgreen}{+0.5})\\
		\cline{2-5}
		& MoCov2 & \multirow{2}{*}{800}   & 65.0 \pt\pt\pt\pt\pt&66.3 \pt\pt\pt\pt\pt\\
		&MoCov2-Tobias&&66.2 (\textcolor{grassgreen}{+1.2})&67.7 (\textcolor{grassgreen}{+1.4})\\
		\cline{2-5}
		& SimCLR & \multirow{2}{*}{200}     &63.6 \pt\pt\pt\pt\pt &64.5 \pt\pt\pt\pt\pt \\
		&SimCLR-Tobias&&65.4 (\textcolor{grassgreen}{+1.8}) & 68.6 (\textcolor{grassgreen}{+4.1})\\
		\cline{2-5}
		& SimCLR & \multirow{2}{*}{800}   & 66.0 \pt\pt\pt\pt\pt&67.3 \pt\pt\pt\pt\pt\\
		&SimCLR-Tobias&&67.4 (\textcolor{grassgreen}{+1.4}) &69.3 (\textcolor{grassgreen}{+2.0})\\
		\hline
		
		\multirow{10}{*}{ResNet-50}& ImageNet super.&N/A &    81.3 \pt\pt\pt\pt\pt&82.1 \pt\pt\pt\pt\pt\\
		\cline{2-5}
		& random init. &0& 58.6 \pt\pt\pt\pt\pt&56.3 \pt\pt\pt\pt\pt\\
		\cline{2-5}
		& MoCov2 & \multirow{2}{*}{200}     &56.2 \pt\pt\pt\pt\pt &53.0 \pt\pt\pt\pt\pt \\
		&MoCov2-Tobias&&63.6 (\textcolor{grassgreen}{+7.4})&62.0 (\textcolor{grassgreen}{+9.0})\\
		\cline{2-5}
		& MoCov2 & \multirow{2}{*}{800}   & 66.5 \pt\pt\pt\pt\pt&62.0 \pt\pt\pt\pt\pt\\
		&MoCov2-Tobias&&67.2 (\textcolor{grassgreen}{+0.7})&71.5 (\textcolor{grassgreen}{+9.5})\\
		\cline{2-5}
		& SimCLR & \multirow{2}{*}{200}     &68.0 \pt\pt\pt\pt\pt&66.5 \pt\pt\pt\pt\pt\\
		&SimCLR-Tobias&&68.4 (\textcolor{grassgreen}{+0.4})&71.7 (\textcolor{grassgreen}{+5.2}) \\
		\cline{2-5}
		& SimCLR & \multirow{2}{*}{800}   & 69.2 \pt\pt\pt\pt\pt &73.0 \pt\pt\pt\pt\pt\\
		&SimCLR-Tobias&&70.0 (\textcolor{grassgreen}{+0.8})& 73.6 (\textcolor{grassgreen}{+0.6})\\
		\hline
	\end{tabular}
\end{table}

We carefully study our Tobias using 2 typical SSL methods, namely MoCov2~\citep{mocov2:xinlei:arxiv2020} and SimCLR~\citep{simclr:hinton:ICML20} under both ResNet-18 and ResNet-50. We follow the training and evaluation protocols in~\citep{S3L:cao:arxiv2021} and conduct experiments on CUB-200. The full learning process contains two stages: pretraining and fine-tuning. We use the pretrained weights obtained by SSL for initialization and then fine-tune networks for classification. Note that \textit{SSL pretraining and fine-tuning are both performed only on the target dataset CUB-200} in this subsection.

For the fine-tuning stage, we fine-tune all methods for 120 epochs using SGD with a batch size of 64, a momentum of 0.9 and a weight decay of 5e-4 for fair comparison. The learning rate starts from 0.1 with cosine learning rate decay. We also list the results using the Mixup strategy, where the alpha is set to 1.0. For the SSL pretraining stage, we follow the same settings in the original papers. `-Tobias' denotes our method and we only change the data loading process and other training settings remain the same as baseline methods.

The results are shown in Table~\ref{tab:clean-cub200-result}. Tobias has consistent improvements under various backbones, pretraining epochs and SSL algorithms. Taking ResNet-50 as an example, our Tobias achieves $13.2\%$ relative higher accuracies than the baseline MoCov2 with normal fine-tuning when both pretrained for 200 epochs. Also, the relative improvement has risen to $17.0\%$ if we use MixUp. It is because that we also merge image patches (in an informative way) during pretraining and it is more friendly to subsequent fine-tuning with MixUp. Moreover, we can observe that the improvement is larger when pretrained for fewer epochs (200 vs. 800). It is because that our method can better capture foreground objects, which leads to faster convergence during pretraining. We will further demonstrate the effectiveness of such foreground vs. background information.

\subsubsection{Results on ImageNet} \label{sec:imagenet}

\begin{table}
	\setlength{\tabcolsep}{3pt}
	\caption{Object detection on PASCAL VOC trainval07+12 (default VOC metric $\text{AP}_{50}$, COCO-style AP, and $\text{AP}_{75}$).}
	\label{tab:coco-result}
	\centering
	\footnotesize
	\begin{tabular}{l|c c c|c c c}
		\multirow{2}{*}{pretraining method}&\multicolumn{3}{c|}{R-50-FPN (24k)} & \multicolumn{3}{c}{R-50 C4 (24k)}\\
		\cline{2-7}
		&$\text{AP}_{50}$ & $\text{AP}$ & $\text{AP}_{75}$ &$\text{AP}_{50}$ & $\text{AP}$ & $\text{AP}_{75}$ \\
		\cline{1-2} 
		\hline
		random init.&63.0&36.7&36.9&60.2&33.8&33.1\\
		IN supervised&80.8&53.5&58.4&81.3&53.5&58.8\\
		\hline
		MoCov2 200ep&81.8&55.0&60.5&82.2&57.1&64.5\\
		
		MoCov2-Tobias 200ep&\textbf{82.0}&\textbf{55.5} &\textbf{61.1} &\textbf{82.6} & \textbf{57.7}&\textbf{64.9} \\
		\hline
		\textcolor{gray}{MoCov2 800ep}&\textcolor{gray}{81.5}&\textcolor{gray}{55.0}&\textcolor{gray}{61.0}&\textcolor{gray}{82.6}&\textcolor{gray}{57.7}&\textcolor{gray}{64.5}\\
	\end{tabular}
\end{table}

Now we move on to the large-scale dataset ImageNet. We use MoCv2 for illustration following the official training protocols in~\citep{mocov2:xinlei:arxiv2020}. We adopt ResNet-50 as backbone and set the batch size to 256, learning rate to 0.03 and number of epochs to 200. We study the downstream object detection performance on Pascal VOC 07\&12~\citep{VOC:mark:IJCV10} in Table~\ref{tab:coco-result}. The detector is Faster R-CNN with a backbone of R-50-FPN~\citep{FPN:kaiming:CVPR17} or R-50-C4~\citep{mask-rcnn:he:ICCV17}, implemented in~\citep{wu2019detectron2}.

As shown in Table~\ref{tab:coco-result}, Tobias achieves better performance than baseline MoCov2 on Pascal VOC. Also notice that our Tobias 200ep even achieves slightly better performance than MoCov2 800ep (pretrained much longer).

\begin{table}
	\caption{Downstream object detection performance on VOC 07\&12 and linear evaluation accuracy on Tiny-IN-200 when pretrained on ImageNet subsets using ResNet-50. `\#imgs' (`\#eps') represent the number of images (epochs).}
	\label{tab:small-imagenet-result}
	\centering
	\small
	\setlength{\tabcolsep}{1.2pt}
	\renewcommand{\arraystretch}{0.95}
	\renewcommand{\multirowsetup}{\centering}
	\begin{tabular}{l|r|r|c|c|c}
		\hline
		\multicolumn{3}{c|}{pretraining}          &           \multicolumn{2}{c|}{VOC 07\&12}    &\multirow{2}{*}{Tiny-IN-200}         \\ 
		\cline{1-5}
		method & \#imgs&\#eps & $\text{AP}_{50}$ & $\text{AP}_{75}$ &\\
		\hline
		random init.                           &0&   0&     63.0 \pt\pt\pt\pt\pt         &       36.9 \pt\pt\pt\pt\pt        & \pt0.5 \pt\pt\pt\pt\pt \\
		\hline
		MoCov2	&\multirow{2}{*}{10k} &   \multirow{2}{*}{200}  &     71.1 \pt\pt\pt\pt\pt    &  45.8 \pt\pt\pt\pt\pt     &  \pt0.5 \pt\pt\pt\pt\pt   \\
		MoCov2-Tobias&&     &          71.4 (\textcolor{grassgreen}{+0.3})    &    47.0  (\textcolor{grassgreen}{+1.2})           & \pt9.9 (\textcolor{grassgreen}{+9.4})  \\
		\cline{1-6}
		MoCov2	&\multirow{3}{*}{10k} &   \multirow{3}{*}{800}  &     71.6 \pt\pt\pt\pt\pt     &  45.9 \pt\pt\pt\pt\pt     &  23.6 \pt\pt\pt\pt\pt   \\
		MoCov2-Tobias&&     &          73.2 (\textcolor{grassgreen}{+1.6})  &    48.5  (\textcolor{grassgreen}{+2.6})           & 23.9 (\textcolor{grassgreen}{+0.3})  \\
		MoCov2-RM&&     &          72.0\pt$\downarrow$1.2\pt     &    47.4\pt$\downarrow$1.1\pt           &  23.5\pt$\downarrow$0.4\pt   \\
		MoCov2-Mixup&&&70.9\pt$\downarrow$2.3\pt &43.3\pt$\downarrow$5.2\pt &19.3\pt$\downarrow$4.6\pt \\
		\cline{1-6} 
		MoCov2&\multirow{2}{*}{50k} &   \multirow{2}{*}{200}  &          72.2 \pt\pt\pt\pt\pt  &      46.8 \pt\pt\pt\pt\pt              &    26.3 \pt\pt\pt\pt\pt \\
		MoCov2-Tobias& &    &     73.7 (\textcolor{grassgreen}{+1.5})  &  49.2 (\textcolor{grassgreen}{+2.4})     &  26.0 (\textcolor{red}{-0.3})\tiny\pt \\
		\cline{1-6} 
		MoCov2&\multirow{3}{*}{50k} &   \multirow{3}{*}{800}  &          77.5 \pt\pt\pt\pt\pt  &    53.3  \pt\pt\pt\pt\pt              &    37.9 \pt\pt\pt\pt\pt \\
		MoCov2-Tobias& &    &     77.9 (\textcolor{grassgreen}{+0.4})   &        54.9 (\textcolor{grassgreen}{+1.6})     & 40.7 (\textcolor{grassgreen}{+2.8}) \\
		MoCov2-RM&&     &          77.4\pt$\downarrow$0.5\pt     &  53.3\pt$\downarrow$1.6\pt           & 40.1\pt$\downarrow$0.6\pt  \\
		MoCov2-Mixup&&&76.7\pt$\downarrow$1.2\pt &52.4\pt$\downarrow$2.5\pt &38.7\pt$\downarrow$2.0\pt \\
		\cline{1-6} 
		MoCov2&\multirow{2}{*}{100k} &   \multirow{2}{*}{200}  &          76.2 \pt\pt\pt\pt\pt &    51.6  \pt\pt\pt\pt\pt                &  35.3 \pt\pt\pt\pt\pt  \\
		MoCov2-Tobias& &    &     77.5 (\textcolor{grassgreen}{+1.3})    &        53.9 (\textcolor{grassgreen}{+2.3})    &  36.5 (\textcolor{grassgreen}{+1.2}) \\
		\cline{1-6} 
		MoCov2&\multirow{2}{*}{100k} &   \multirow{2}{*}{800}  &          78.7 \pt\pt\pt\pt\pt    &    56.3  \pt\pt\pt\pt\pt                &  43.7 \pt\pt\pt\pt\pt  \\
		MoCov2-Tobias& &    &     79.4 (\textcolor{grassgreen}{+0.7})    &        57.3 (\textcolor{grassgreen}{+1.0})    &  44.3 (\textcolor{grassgreen}{+0.6}) \\
		\hline
	\end{tabular}
\end{table}

Apart from the full large-scale ImageNet dataset, we also study the performance under different data volumes by sampling the original ImageNet to smaller subsets, motivated by~\citet{S3L:cao:arxiv2021}. We randomly sample (\textit{without using any image label}) 10 thousand (10k), 50 thousand (50k) and 100 thousand (100k) images to construct IN-10k, IN-50k and IN-100k, respectively. We only change the amount of data here and other training settings remain the same as before. The results are shown in Table~\ref{tab:small-imagenet-result} and we adopt Pascal VOC 07\&12 for object detection and Tiny-ImageNet-200 (100,000 training and 10,000 validation images from 200 classes at $64 \times 64$ resolution) for linear evaluation. 

\begin{figure}
	\centering
	\includegraphics[width=0.9\linewidth]{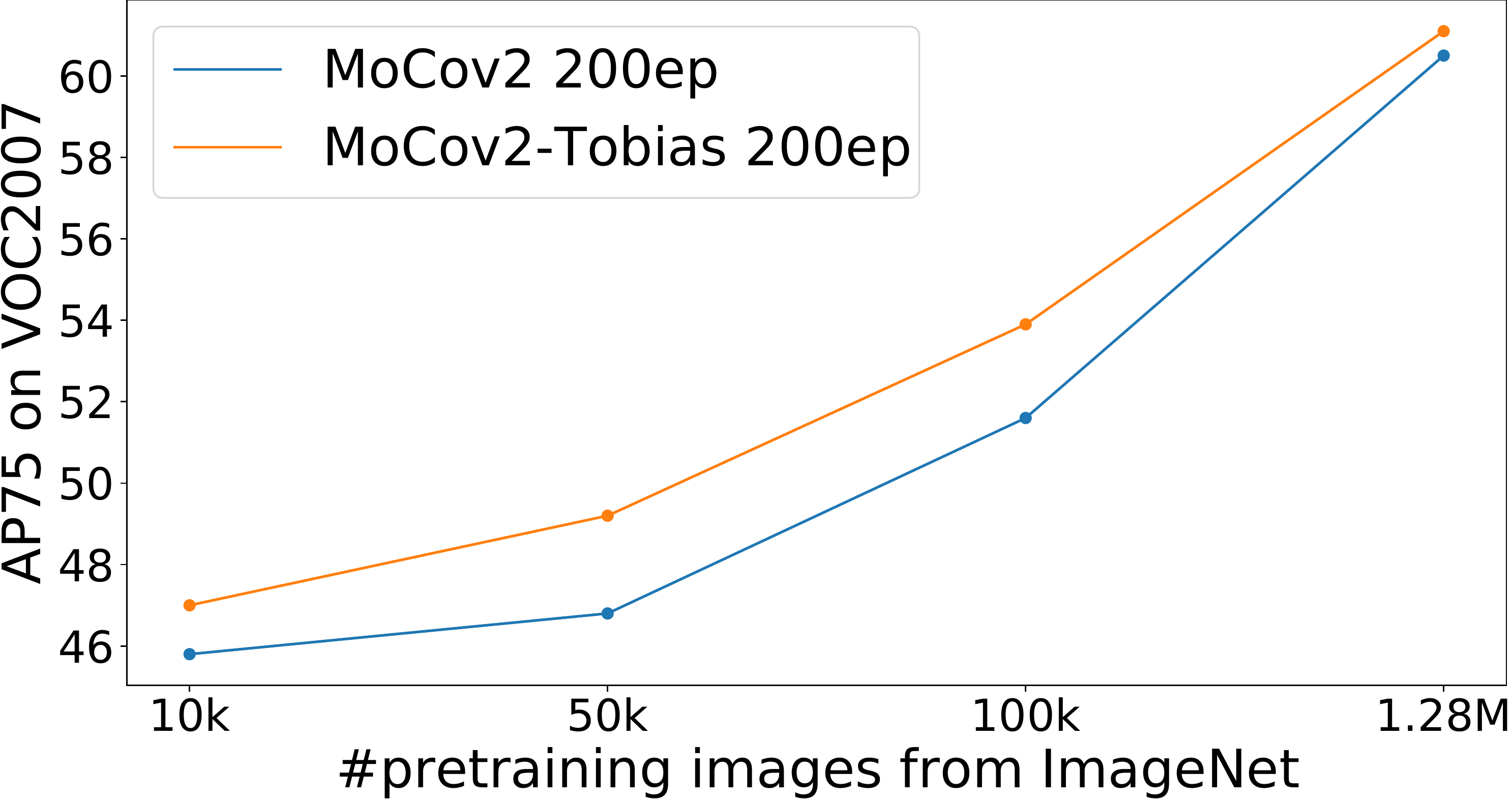}
	\caption{Performance of Tobias on Pascal VOC ($\text{AP}_{75}$) with respect to different training data size.}
	\label{fig:ap-size}
\end{figure}

As can be seen in Table~\ref{tab:small-imagenet-result} and Figure~\ref{fig:ap-size}, our Tobias achieves significant improvements on both downstream tasks, especially on VOC 07\&12 object detection. For instance, when both trained for 200 epochs on IN-100k, our Tobias is significantly better than baseline counterpart: up to \textbf{+1.3} $\text{AP}_{50}$ and \textbf{+2.3} $\text{AP}_{75}$. Also notice that when both trained for 200 epochs on IN-10k, MoCov2 performs the same as random guess (0.5\%) while our method learns much better representations (9.9\%) in terms of Tiny-IN-200 linear evaluation. In general, our method improves the most on $\text{AP}_{75}$, which is a more stringent metric for detection accuracy. It indicates that our model can better capture foreground objects across changing backgrounds during pretraining, hence improving performance for object detection as well as image classification. Moreover, our method is especially effective (i.e., has greater improvements) when the amount of data is small.

\subsection{Ablation studies} \label{sec:exp3}

In this section, we will first study the effectiveness of the foreground vs. background (Tobias) information (generated by random networks). Then, we will study the effect of the hyper-parameter $p$ in our method. Finally, we study the sensitivity to data augmentations.

\textbf{Effect of Tobias information.} Notice that we use the foreground vs. background information when merging patches from two images. To demonstrate its effectiveness, we design a random merging strategy for comparison (MoCov2-RM in Table~\ref{tab:small-imagenet-result}). More specifically, we do not use such information and randomly select patches from two images for merging (also half-half division) and it can also be viewed as one kind of patch-level CutMix. We also compare with MoCov2-Mixup where we use Mixup when merging images. We keep all other settings the same and conduct pretraining on both IN-10k and IN-50k. As can be seen in Table~\ref{tab:small-imagenet-result}, we will see a significant drop, especially in object detection performance if we discard the foreground vs. background information provided by our Tobias: up to \textbf{-1.2} $\text{AP}_{50}$ for RM and \textbf{-2.3} $\text{AP}_{50}$ for Mixup when trained on IN-10k for 800 epochs. It demonstrates the Tobias information provided by a randomly initialized network is vital. Another interesting thing is that RM achieves better performance than the baseline MoCov2, which indicates that this kind of data augmentation is somehow useful for SSL, as shown in ~\citep{unmix:shen:arxiv20}.   

\textbf{Effect of hyper-parameter.} Now we study the effect of the hyper-parameter $p$, i.e., the probability of changing backgrounds in another view. We study $p$ = 0, 0.3, 0.5, 0.7 and 1.0. Notice that when $p$=0, our Tobias degenerates into the baseline MoCov2. We train on IN-10k for 800 epochs for all settings in Table~\ref{tab:hyperparameter}. For object detection, we can see that when $p$ grows, the result becomes better and will not continue to improve when it grows beyond 0.5. For Tiny ImageNet, $p=0.7$ achieves the highest accuracy. Notice that we \emph{directly set $p$ to 0.3 for all our experiments throughout this paper} and did not tune it under different settings. It also indicates that we can get better results with more carefully tuned $p$.

\textbf{Sensitivity to image augmentations.} Now we study the sensitivity to image augmentations of our Tobias by progressively removing transformations in the transformation set following~\citet{byol:grill:NIPS20}. The results in Table~\ref{tab:transformation} show that the performance of Tobias is much less affected than the performance of MoCov2 when removing the color distortion from the set of image augmentations, especially on Tiny-IN-200. Also we can observe that color distortion (e.g., grayscale and color-jitter) has greater impact on downstream image classification and less impact on object detection. When image augmentations are reduced to a mere random crop, the gap between our Tobias and baseline MoCov2 has increased to 2.9 and 3.2 points for Tiny-IN-200 and VOC detection ($\text{AP}_{75}$), respectively. It indicates that our Tobias is itself an effective data augmentation and less sensitive to other augmentations.

\begin{table}
	\caption{Effect of hyper-parameter $p$. All settings are pretrained on IN-10k for 800 epochs using ResNet-50.}
	\label{tab:hyperparameter}
	\centering
	\small
	\renewcommand{\arraystretch}{0.9}
	\renewcommand{\multirowsetup}{\centering}
	\begin{tabular}{c|ccc|c}
		\hline
		\multirow{2}{*}{prob $p$}         &           \multicolumn{3}{c|}{VOC 07\&12}    &\multirow{2}{*}{Tiny-IN-200}         \\ 
		\cline{2-4}
		& $\text{AP}_{50}$ & $\text{AP}$ & $\text{AP}_{75}$ &\\
		\hline
		0.0	 &     71.6      &  43.9       &  45.9      &  23.6    \\
		0.3	 &     73.2     &  45.7       &  48.5      &  23.9    \\
		0.5	 &     \textbf{73.9}     &  \textbf{46.3}       &  \textbf{49.4}      &  23.3    \\
		0.7	 &     72.3     &  44.8       &  47.4      &  \textbf{25.4}    \\
		1.0 &       71.8      & 44.3&46.6&24.3\\
		\hline
	\end{tabular}
\end{table}

\begin{table}
	\caption{Impact of progressively removing transformations. All pretrained on IN-10k for 800 epochs.}
	\label{tab:transformation}
	\centering
	\small
	\setlength{\tabcolsep}{2pt}
	\renewcommand{\arraystretch}{0.95}
	\renewcommand{\multirowsetup}{\centering}
	\begin{tabular}{c|ccc|ccc}
		\hline
		\multirow{2}{*}{transformation set}         &           \multicolumn{3}{c|}{MoCov2}    &\multicolumn{3}{c}{MoCov2-Tobias}         \\ 
		\cline{2-7}
		& $\text{AP}_{50}$ &  $\text{AP}_{75}$ &Tiny-IN & $\text{AP}_{50}$ & $\text{AP}_{75}$ &Tiny-IN\\
		\hline
		baseline	 &     71.6          &   45.9      &   23.6\tiny{\pt\pt\pt\pt\pt} &73.2&48.5&23.9\tiny{\pt\pt\pt\pt\pt} \\
		remove grayscale	 &   70.2          &   44.1    &      19.9\tiny$\downarrow3.7$&73.1&49.0&22.7\tiny$\downarrow1.2$\\
		remove color 	 &   71.3            &  46.0       &18.1\tiny$\downarrow5.5$   &72.7&48.2&21.2\tiny$\downarrow2.7$ \\
		crop+flip only &    71.0   &46.2 & 16.8\tiny$\downarrow6.8$&72.9&48.3& 20.2\tiny$\downarrow3.7$\\
		crop only &71.7&46.7&15.0\tiny$\downarrow8.6$&73.1&49.9&17.9\tiny$\downarrow6.0$\\
		\hline
	\end{tabular}
\end{table}

\section{Conclusions}

In this paper, we revealed the phenomenon that a randomly initialized CNN has the potential to localize objects well, which we called Tobias. Moreover, we analyzed that activation functions like ReLU and network depth are essential for a random CNN to localize. Then, we proposed Tobias self-supervised learning, which forces the model to focus on foreground objects by dynamically changing backgrounds while keeping the objects under the guidance of Tobias. Various experiments have shown that our method obtained a significant edge over baseline counterparts because it learns to better capture foreground objects. In the future, we will try to apply our Tobias to supervised learning. 

\bibliography{egbib}

\end{document}